\documentclass[runningheads]{llncs}
\usepackage{graphicx}
\begin{document}
\title{Joint Extraction and Classification of Danish Competences for Job Matching}
\titlerunning{Danish Job Competence Extraction and Classification}
%
\author{Qiuchi Li\orcidID{0000-0002-8219-0869} \and
Christina Lioma\orcidID{0000-0003-2600-2701}}
\authorrunning{Q. Li and C. Lioma}
%
\institute{University of Copenhagen \\ Universitetsparken 1, 2100 Copenhagen, Denmark \\
\email{\{qiuchi.li, c.lioma\}@di.ku.dk}}
\maketitle              
\begin{abstract}
The matching of competences, such as skills, occupations or knowledges, is a key desiderata for candidates to be fit for jobs. Automatic extraction of competences from CVs and Jobs can greatly promote recruiters' productivity in locating relevant candidates for job vacancies. This work presents the first model that jointly extracts and classifies competence from Danish job postings. Different from existing works on skill extraction and skill classification, our model is trained on a large volume of annotated Danish corpora and is capable of extracting a wide range of danish competences, including skills, occupations and knowledges of different categories. More importantly, as a single BERT-like architecture for joint extraction and classification, our model is lightweight and efficient at inference. On a real-scenario job matching dataset, our model beats the state-of-the-art models in the overall performance of Danish competence extraction and classification, and saves over 50\% time at inference.

\end{abstract}

\section{Introduction}

Job matching, also known as person-job fit or job-resume matching, is a crucial and challenging scenario in job recruitment where matchers need to search suitable candidates for job vacancies from a huge pool of candidate profiles. The booming increase in job vacancies on recruitment platforms creates a high demand for prompt and accurate identification of matched candidates, placing a great burden to recruiters~\cite{montuschi_job_2014}. The absence of such matching systems will cause financial losses to both job seekers and companies~\cite{cardoso_matching_2021}.

Competences, specifically skills, knowledges or occupations, serve as one of the most important criteria for judging the relevant candidates~\cite{bogers_exploration_2021}. Accurate and prompt extraction of competences from Jobs and CVs can promote accurate matching of relevant candidates and liberate recruiters' from their burden. While existing works~\cite{zhu_person-job_2018,dave_combined_2018,yan_interview_2019,qin_enhanced_2020,LiLioma2022} have mainly attempted to match job and candidates by their representation as a whole, the extraction of competences can support a clear presentation of reasons of matching along with the matching result. However, the job matching context poses greater challenges to competence extraction algorithms in the following aspects: 
\begin{itemize}
    \item \textbf{Extraction accuracy.} A recruiter is often not knowledgeable to the industry related to the matching task, and relies on the extracted competences for finding the relevant candidates. Therefore, the extracted competence should be of high quality to support real matching scenario.

    \item \textbf{Fine-grained categories.} A competence can be expressed in different ways. Apart from exact term matching, fine-grained categorization of extracted competences should be devised to account for this issue and promote finding more relevant candidates.

    \item \textbf{Efficiency.}  For productivity, a competence extraction model should be able to generate prompt response to incoming jobs and CVs.
    
\end{itemize}

Machine learning algorithms have been developed for automatic identification~\cite{jia_representation_2018,sayfullina_2018_learning}, extraction~\cite{Gugnani_Misra_2020,zhang-2022-skillspan,tamburri_2020_dataops,chernova2020occupational} and classification~\cite{zhang-2022-kompetencer} of competences from Job postings or CVs. These works mainly target English or Chinese job postings. For Danish Jobs, however, the research on competence extraction is limited by the lack of available annotated data. Zhang et al.~\cite{zhang-2022-kompetencer} investigated Danish competence classification with distant supervision. On their collected tiny-scale dataset of 60 Danish job postings, few-shot learning and cross-language transfer learning led to decent performance. However, the extraction of Danish competences is still an unsolved task. 

We frame the task of Danish competence extraction as a token classification task, and propose a novel model for jointly extracting and classifying Danish competences for job matching. The model is based on pre-trained text encoder for Danish job postings~\cite{zhang-2022-skillspan} and maps the encoded sentence representations to produce named entity recognition (NER) labels for competence extraction and multi-class labels for competence classification. The model is trained on around 200,000 sentences from Danish Jobs and CVs with annotations of European Skills, Competences, Qualifications and Occupations (ESCO)~\cite{le_2014_esco}. Different from~\cite{zhang-2022-kompetencer}, we include a wider range of competences in the ESCO taxonomy, broadly covering the main categories of skills, occupations and knowledges. The model is jointly trained from annotation labels of both tasks, and extracts and classifies competences in separate steps at prediction. Our model achieves 
 improved accuracy over the best existing practices on fine-grained competence extraction and classification, and takes only half of the prediction time.\\


\section{Task Definition and Data Description}

The task is to extract and classify Danish Competences from job postings. Specifically, the input is a Danish sentence or a sequence of tokens $\mathcal{X} = \{x_1, x_2,...,x_N\}$, and the output is a list of (text span, class) tuples $\mathcal{Y} = \{(s_1, c_1), (s_2, c_2),..., (s_K, c_K)\}$. Each span $s_k = \{x_i\}_{i=k_{start}}^{k_{end}}$ is a continuous sub-sequence of tokens of $\mathcal{X}$, and $c_k \in \mathcal{C}$ is the class label of $s_k$ that belongs to a pre-set collection of labels.  Under this notation, SKILLSPAN~\cite{zhang-2022-skillspan} targets at establishing the mapping from $\mathcal{X}$ to $\mathcal{Y^{S}} = \{s_1, s_2, ..., s_K\}$, while KOMPETENCER~\cite{zhang-2022-kompetencer} manages to predict $c_k$ for each input $s_k$. We seek to directly learn the mapping $\mathcal{X} \rightarrow \mathcal{Y}$ from annotated data with a single model.

We proposed to jointly extract and classify Danish competences. 
The extraction of competences entities is formulated as a named entity recognition (NER) task, where a 3-class label is predicted for each token: [\textbf{O, I, B}]. \textbf{B} marks the beginning of an entity, \textbf{I} refers to the inner part of an entity, and \textbf{O} stands for a non-entity token. A ``\textbf{BII...I}'' pattern indicates a multi-token entity. The classification of entities is formulated as a multi-class classification task. 

An important ingredient to our model is the resource for Danish competences. For this purpose, we rely on the European Skills, Competences, Qualifications and Occupations (ESCO) taxonomy, which contains a total number of 16898 skills, occupations, knowledges in 28 different languages. Each ESCO entity has a textual description and associated to 4-leveled annotations. This work aims at extracting text spans that are considered as ESCO entities, and further classify them into top-level categories in the ESCO taxonomy, include 10 occupation categories ($C_0-C_9$), 8 skill categories ($S_1-S_8$), 2 language skill categories ($L_0, L_1$), 11 knowledge categories ($K_{00}-K_{10}$), as well as 6 transversal skills and competences ($T_1-T_6$). We also include three labels ($C_{-1}, K_{-1}, S_{-1}$) for non-ESCO occupations, knowledges and skills. 

\begin{figure*}[t]
    \centering
    \includegraphics[width=\textwidth]{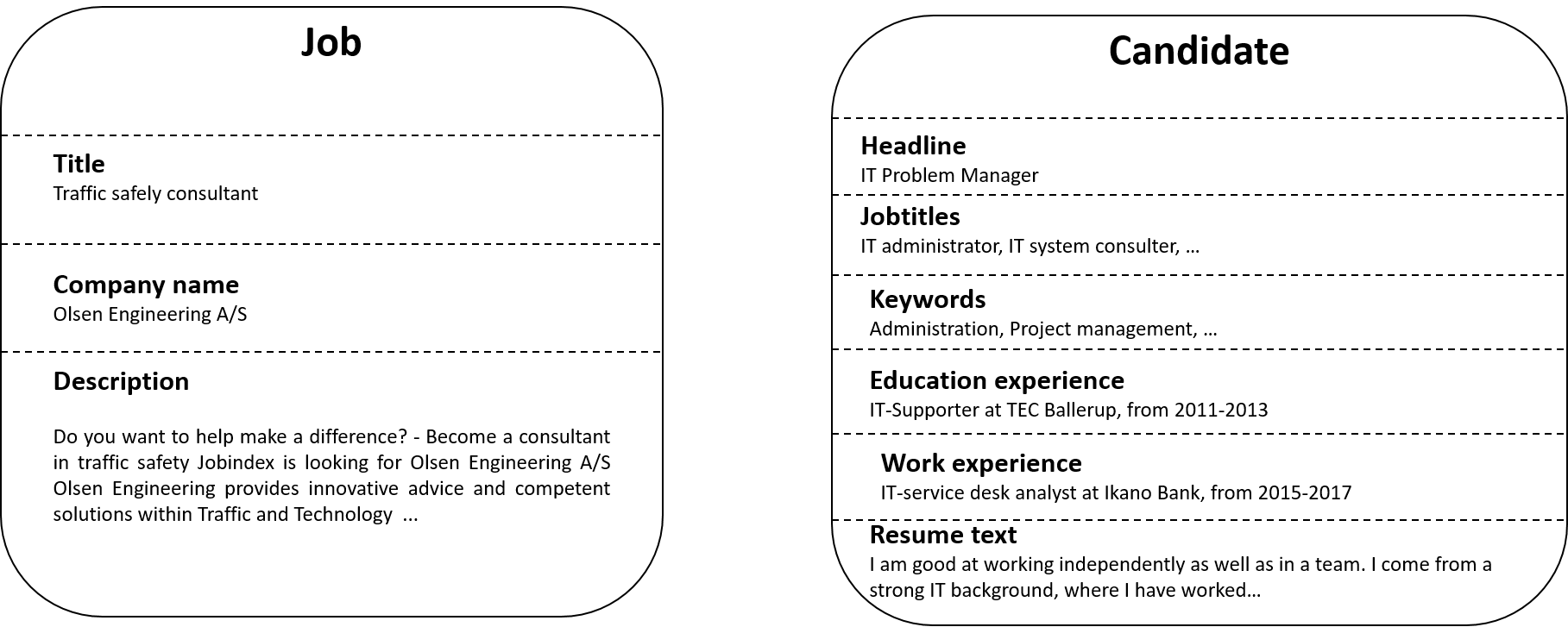}
    \caption{Ingredients of a job posting and a candidate profile in the Jobindex database. The texts are translated to English for a better understanding.}
    \label{fig:example}
\end{figure*}

We apply our model on a collection of annotated sentences from the Jobindex\footnote{https://www.jobindex.dk/} database. Jobindex is a job portal located in Denmark. It originally targeted at the Danish market and has expanded to have sites in 3 other countries. The sentences come from an abundance of Danish jobs and candidate profiles - see Fig.~\ref{fig:example} for an illustration of the main text fields in each of them. For jobs, the sentences come from its textual descriptions. From candidate profiles, we extract sentences from educational and work experience. We apply exact phrase matching to detect ESCO entities in each sentence, split the sentence into a sequence of tokens, and insert NER labels and class labels based on the extracted ESCO entities. 

\section{Our Model}
\begin{figure*}[t]
    \centering
    \includegraphics[width=\textwidth]{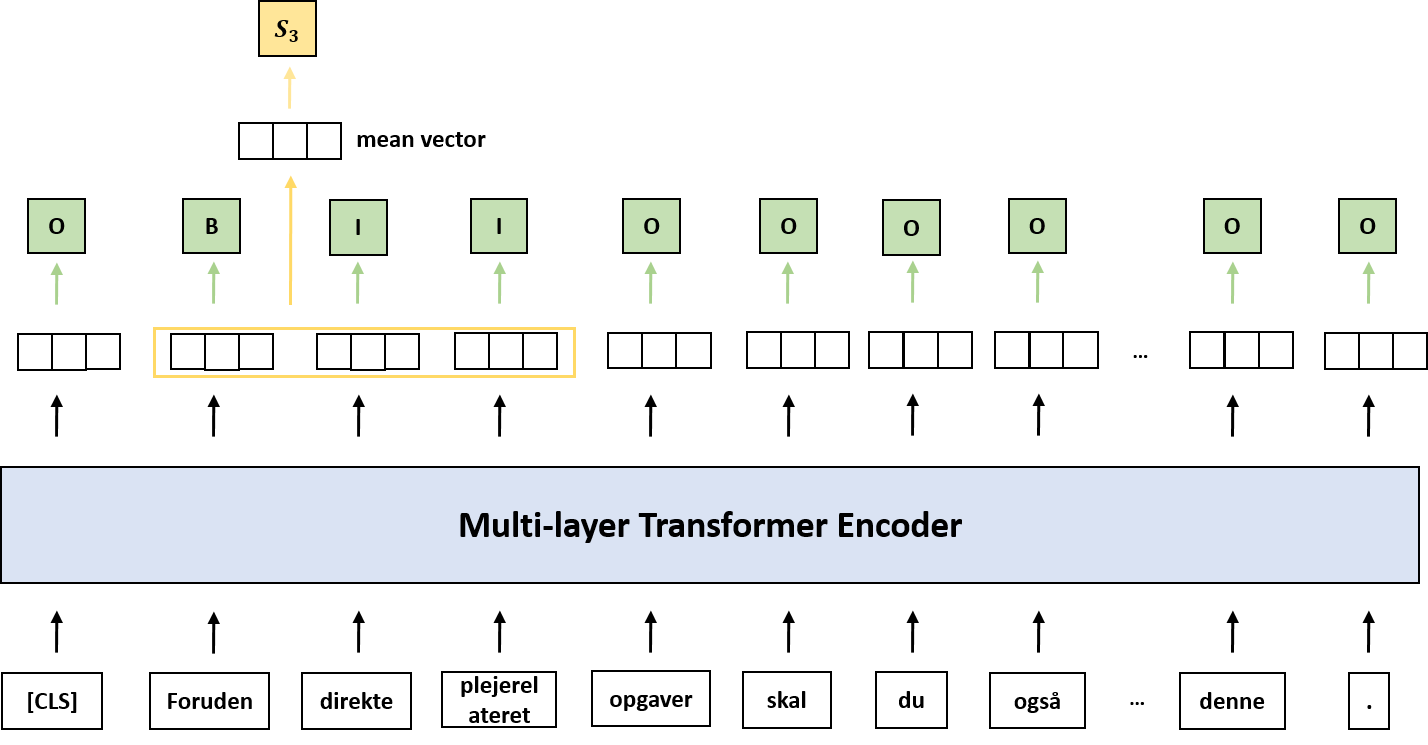}
    \caption{The architecture of the joint ESCO extraction and classification model.}
    \label{fig:model}
\end{figure*}

We build a single model to tackle ESCO entity extraction and classification. As shown in Fig.~\ref{fig:model}, NER labels and ESCO class labels are produced based on a multi-layer Transformer text encoder. The model jointly learns from annotated labels of both tasks in the training step, but produces the labels in a sequential manner at prediction.

\subsection{Text Encoder}
We take existing Danish BERT models for the text encoder, and formulate the extraction and classification of ESCO labels as a fine-tuning task. Danish BERT (\textbf{DaBERT})\footnote{https://huggingface.co/Maltehb/danish-bert-botxo} is a publicly available BERT model trained by Certainly\footnote{https://certainly.io/} on 9.5GB Danish texts. Zhang et al.~\cite{zhang-2022-kompetencer} obtained a Danish BERT for job recommendation context, namely \textbf{DaJobBERT}, by further training \textbf{DaBERT} on 24.5M Danish job posting sentences for one epoch. \textbf{DaJobBERT} is reported to have superior few-shot performance over \textbf{DaBERT}~\cite{zhang-2022-kompetencer}. Both encoders are included in this work.
 
\subsection{ESCO Detection and Classification} 

Token-wise NER labels are produced for detecting ESCO entities. We construct simple feed-forward neural network $f_{NER}$, containing a single hidden layer with Tanh as the activation function, for mapping each encoded token vector to a 3-class NER label. 

For ESCO classification, all tokens in the extracted ESCO entities are aggregated to a fixed-dimensional vector. Specfically, we take the average of the encoded token vectors for each entity span, and pass it to a feed-forward network $f_{CLS}$ to produce its ESCO class label. 

\subsection{Joint Training}
A training sample is annotated with NER labels for all tokens and ESCO class labels for the ESCO entities. ESCO detection loss $\mathcal{L}_{NER}$ is the average cross-entropy loss over all tokens against the golden NER labels in a sentence. For ESCO classification loss, we pass the average token vectors of each \textbf{golden entity} to $f_{CLS}$. and compute the average cross-entropy loss between the output logits and true ESCO labels over all entities as the ESCO classification loss $\mathcal{L}_{CLS}$.  The overall training loss is a combination of both losses controlled by a linear weight $\lambda$:

\begin{equation}
    \mathcal{L} = \mathcal{L}_{CLS} + \lambda  \mathcal{L}_{NER}.
\end{equation}

Both the text encoder and feed-forward networks ($f_{CLS}, f_{NER}$) are learned by minimizing the loss $\mathcal{L}$ with a standard back-propagation algorithm.

\subsection{Two-step Prediction} 
The model extracts and classifies ESCO entities in separate steps for an input sentence at prediction. First, the sentence is passed to the text encoder and NER network $f_{NER}$ to compute an NER label for each token. The entities are extracted accordingly: we detect the \textbf{B} labels in the tokens, and at take the longest \textbf{BII...I} sequence as the entities for each \textbf{B} label. Then, the encoded tokens for each extracted entity are passed $f_{CLS}$ to produce ESCO class labels.

\section{Experiment}
\noindent \textbf{Data.} We evaluate different models for ESCO extraction and classification on job and candidate texts in the Jobindex database. A total number of 217661 sentences are obtained. We split the data into training, validation and test sets at a ratio of 8:1:1.\\

\noindent \textbf{Models.} In addition to our model, a list of BERT-based models are included. Due to the absence of existing models for the same purpose, we aggregate SKILLSPAN~\cite{zhang-2022-skillspan} and KOMPETENCER~\cite{zhang-2022-kompetencer} into a two-step pipeline as the state-of-the-art model. As another two-step approach, we train a separate model based on the competence extraction and classification architecture part of our model. The idea is to check if jointly rendering two tasks leads to improved model capacity. Furthermore, we include an intuitive single-model strategy that views the joint extraction and classification of ESCO competences as an end-to-end NER model, where class-specific NER labels are produced, such as \textbf{B-S$_1$}, \textbf{I-L$_0$}, etc. All models above are in Danish and use the same \textbf{DaJobBERT} checkpoint as the text encoder. We also train our model based on \textbf{DaBERT}, the general-purpose Danish BERT model. By doing so we aim at examining whether a domain-adapted Danish BERT can further enhance our model capacity. 

All models have a 12-layer, 12-attention-head structure with a model size of 768 and an intermediate size of 3072. We use AdamW~\cite{adamw} as the optimizer. We train on the training dataset for one epoch and do early stopping in terms of the validation loss. We set $\lambda=0.1$ to place more attention of our model on competence classification. All models are implemented in the Hugging Face toolkit\footnote{https://huggingface.co/} under PyTorch 1.9.0 and trained on a GPU server with 4 Tesla A100 40GB cards.\\

\noindent \textbf{Metrics.} The same effectiveness metrics is applied to all models. For competence extraction, we compute the F1 score for judging whether a token belongs to an ESCO entity. For competence classification, the weighted macro-F1 score is computed following the established practice~\cite{zhang-2022-kompetencer}. It is a weighted average of F1 scores for all classes considering the support for each class. To check the model efficiency in a real job matching scenario, we also test the average processing time on each job, over the same set of 100 random job postings.

\section{Results and Discussions}

\begin{table}[ht]
\centering
\caption{Size, inference time and effectiveness metrics for models in the experiment. For effectiveness metrics, we compute the average and standard deviation of each value over 5 runs of different random seeds. }
\label{tab:main}
\resizebox{\textwidth}{!}{
\begin{tabular}{c|c|c|c|c}
\hline\hline
\begin{tabular}[c]{@{}c@{}}Model \\Name\end{tabular}& \begin{tabular}[c]{@{}c@{}}Model \\ Size\end{tabular}& \begin{tabular}[c]{@{}c@{}}Inference \\ Time\end{tabular} & \begin{tabular}[c]{@{}c@{}}ESCO Extraction\\ (F1)\end{tabular}& \begin{tabular}[c]{@{}c@{}}ESCO Classification\\ (Weighted Macro F1)\end{tabular} \\ \hline
\begin{tabular}[c]{@{}c@{}}SKILLSPAN + \\ KOMPETENCER\end{tabular} & 862.12M & 1.72s  & \multicolumn{1}{c|}{0.864$\pm$ 0.002} &   \multicolumn{1}{c}{0.436$\pm$ 0.003}   \\ \hline
\begin{tabular}[c]{@{}c@{}}Detection + \\Classification\end{tabular} & 858.39M & 2.14s &\multicolumn{1}{c|}{0.865 $\pm$ 0.003} &  \multicolumn{1}{c}{0.337 $\pm$ 0.002}  \\ \hline
\begin{tabular}[c]{@{}c@{}}All-class NER \end{tabular} & 429.35M & 0.67s & \multicolumn{1}{c|}{0.856 $\pm$ 0.001 } &  \multicolumn{1}{c}{0.588 $\pm$ 0.000 }  \\ \hline
\begin{tabular}[c]{@{}c@{}}Our Model - \\DaJobBERT\end{tabular}  & 431.54M & 0.73s & \multicolumn{1}{c|}{0.860 $\pm$ 0.001} &  \multicolumn{1}{c}{0.623 $\pm$ 0.001 }   \\ \hline
\begin{tabular}[c]{@{}c@{}}Our Model - \\DaBERT\end{tabular} & 431.54M & 0.70s & \multicolumn{1}{c|}{0.841 $\pm$ 0.000} &  \multicolumn{1}{c}{0.627 $\pm$ 0.003}   \\ \hline\hline
\end{tabular}
}
\end{table}

\noindent \textbf{Overall performance.} As shown in Tab.~\ref{tab:main}, all models have similar capacity in extracting ESCO entities. In comparison, SKILLSPAN achieved an F1 score between 0.55 and 0.65 on a relatively small dataset of around 15000 English job sentences~\cite{zhang-2022-skillspan}. The relatively high ESCO extraction performance proves that large-scale annotated data contributes to strong ESCO extraction capabilities. For classification of ESCO labels, however, our model beats the existing works by a remarkable margin. In addition, the SOTA solution takes twice as much storage, and takes slightly more than twice as much as our model due to the intermediate data processing steps between the two models. All the above advantages makes our model an obviously better candidate under the job matching scenario. \\

\noindent \textbf{Effect of joint learning.} The huge gap in the competence classification performance between single-model and two-model approaches reveals a huge positive influence of ESCO extraction to ESCO classification, which could be well captured by the joint learning architecture. Compared to a single NER prediction task for all classes, our structure is a better proposal. Since the all-class NER view leads to a doubling of class numbers, the class imbalance and data sparsity issue are likely to bring negative impact to competence classification.\\

\noindent \textbf{Effect of domain-adapted Danish BERT.} Building our model based on \textbf{DaBERT} encoder yields close performance to on the \textbf{DaJobBERT} encoder. This is to the contrary of the observations in~\cite{zhang-2022-kompetencer}, where \textbf{DaJobBERT} significantly outperformed \textbf{DaBERT} in the few-shot setting. We have demonstrated that, with sufficient in-domain data, different pre-trained Danish BERT encoders have minimal influence to competence extraction and classification.  \\ 
\section{Conclusion}

We present a novel model for jointly extracting and classification Danish competences for Job Matching. On a large collection of annotated samples, this model excels at extracting competences of fine-grained categories, in over 50\% less time compared to the SOTA approach. The strong effectiveness and efficiency makes it better at tackling the requirements of job matching. 

This work is limited to Danish language and the model is not evaluated on a publicly available dataset. We will examine if similar findings hold on publicly available English job postings. We also plan to integrate this model to an automatic job recommendation framework to directly study its impact on a real job matching scenario.\\

\noindent \textbf{Acknowledgement.} This research was supported by the Innovation Fund Denmark, grant no. 0175-000005B. We are grateful for Jobindex’s support on providing the data and setting up the experiment.
\bibliographystyle{splncs04}
\bibliography{ecir}
\end{document}